\documentclass[10pt,journal]{IEEEtran}
\usepackage{amsmath,amsfonts}
\usepackage{algorithmic}
\usepackage{algorithm}
\usepackage{array}
\usepackage[caption=false,font=normalsize,labelfont=sf,textfont=sf]{subfig}
\usepackage{textcomp}
\usepackage{stfloats}
\usepackage{url}
\usepackage{verbatim}
\usepackage{graphicx}
\usepackage{cite}
\usepackage{balance}

\usepackage{amsthm}
\usepackage{amssymb}
\usepackage{booktabs}
\usepackage{multirow}
\usepackage{xcolor}
\usepackage{float}
\usepackage{tikz}
\usepackage{tkz-graph}

\usepackage[switch]{lineno} 

\usetikzlibrary{arrows, decorations.markings}

\tikzstyle{vecArrow} = [thick, decoration={markings,mark=at position
   1 with {\arrow[semithick]{open triangle 60}}},
   double distance=2pt, shorten >= 5.5pt,
   preaction = {decorate},
   postaction = {draw,line width=2pt, white,shorten >= 4.5pt}]
\tikzstyle{innerWhite} = [semithick, white,line width=2pt, shorten >= 4.5pt]

\DeclareMathOperator*{\argmin}{arg\,min}
\makeatletter
\DeclareRobustCommand{\sqcdot}{\mathbin{\mathpalette\morphic@sqcdot\relax}}
\newcommand{\morphic@sqcdot}[2]{%
  \sbox\z@{$\m@th#1\centerdot$}%
  \ht\z@=.33333\ht\z@
  \vcenter{\box\z@}%
}

\begin{document}


\title{Graph Portfolio: High-Frequency Factor Predictor via Heterogeneous Continual GNNs}

\author{Min Hu, Zhizhong Tan, Bin Liu~\IEEEmembership{Member,~IEEE,} Guosheng Yin~\IEEEmembership{Senior Member,~IEEE}

\thanks{Min Hu is with School of Finance, Southwestern University of Finance and Economics, China (E-mail: min\_hu@smail.swufe.edu.cn). Zhizhong Tan is with School of Computer Science and Engineering, Faculty of Innovation Engineering, Macau University of Science and Technology, China (E-mail: zhizhongtan@163.com). Bin Liu is with the Center of Statistical Research, School of Statistics, Southwestern University of Finance and Economics, China (Corresponding author, E-mail: liubin@swufe.edu.cn). Guosheng Yin is with Department of Statistics and Actuarial Science, The University of Hong Kong, Hong Kong, China (E-mail: gyin@hku.hk).\\ 
}}

\markboth{IEEE Transactions on xxxx, ~Vol.~xx, No.~xx, xx~xxxx}%
{Shell \MakeLowercase{\textit{et al.}}: High-Frequency Factor Preditor via Heterogeneous Continual Graph Neural Network}


\maketitle

\begin{abstract}
This study aims to address the challenges of futures price prediction in high-frequency trading (HFT) by proposing a continuous learning factor predictor based on graph neural networks.
The model integrates multi-factor pricing theories with real-time market dynamics, effectively bypassing the limitations of existing methods that lack financial theory guidance and ignore various trend signals and their interactions. We propose three heterogeneous tasks, including price moving average regression, price gap regression and change-point detection to trace the short-, intermediate-, and long-term trend factors present in the data. In addition, this study also considers the cross-sectional correlation characteristics of future contracts, where prices of different futures often show strong dynamic correlations. Each variable (future contract) depends not only on its historical values (temporal) but also on the observation of other variables (cross-sectional). To capture these dynamic relationships more accurately, we resort to the spatio-temporal graph neural network (STGNN) to enhance the predictive power of the model. The model employs a continuous learning strategy to simultaneously consider these tasks (factors). Additionally, due to the heterogeneity of the tasks, we propose to calculate parameter importance with mutual information between original observations and the extracted features to mitigate the catastrophic forgetting (CF) problem. Empirical tests on 49 commodity futures in China's futures market demonstrate that the proposed model outperforms other state-of-the-art models in terms of prediction accuracy. Not only does this research promote the integration of financial theory and deep learning, but it also provides a scientific basis for actual trading decisions.

\end{abstract}


\section{Introduction}
\IEEEPARstart{H}{igh-frequency} trading (HFT), which greatly benefits from the development and application of computer technologies, can catch trading opportunities in the rapidly changing market and make transactions promptly, reaping high returns. In most markets, the market share of HFT has reached more than 50\%\cite{musciotto2021high}, and even 70\% in the North American market. Hence, HFT has an important impact on the ecology of the international financial market. 
Relying on rapid information processing and high-speed trading execution, the greatest challenge that HFT faces is accurately predicting short-term price fluctuations. This is particularly crucial in the futures market, which has a high leverage effect and significant price fluctuations. But in practice, due to the microstructure noise in financial markets which is sufficient to skew the asset pricing process \cite{ait2005often}, it is extremely harder to predict the short-term trend than the long-term trend.

Futures price prediction is a problem within multi-variate time series forecasting, deep learning breaks through the form of parametric analytical equation of traditional statistical models \cite{sims1980macroeconomics,box2015time,bollerslev1986generalized}, and has made many advancements in recent years\cite{shih2019temporal,wu2020connecting, sen2019think} by learning the internal rules and representation levels of time series data.

\begin{figure}[!t]
    \centering     \includegraphics[width=1.0\columnwidth]{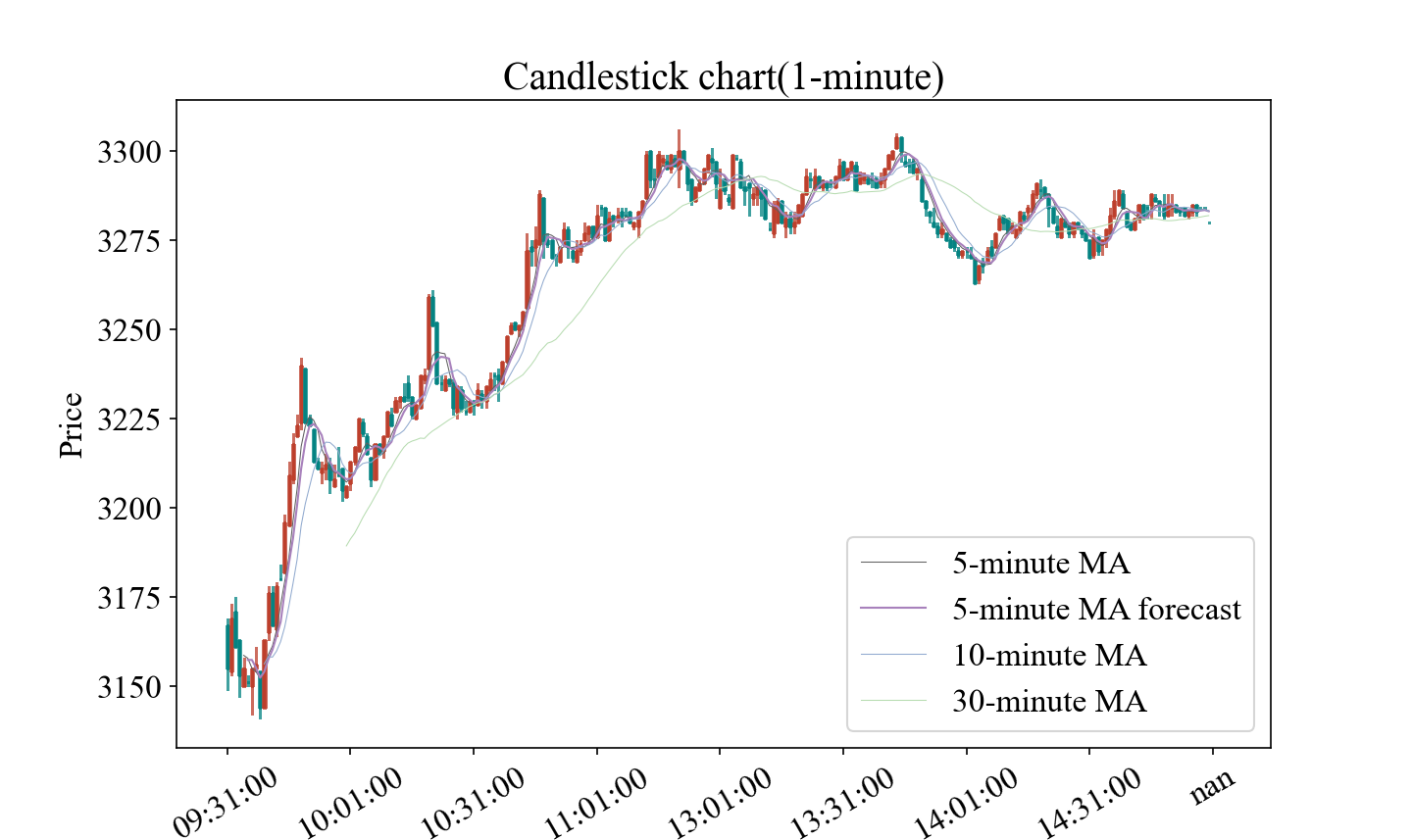}
     \caption{Short-term price signal and the corresponding long-term trends.}
     \label{fig:shortLongTerm}
\end{figure}

More often than not, financial systems are complex and their underlying dynamics are not known \cite{taylor2009composable}. There are at least two significant characteristics of the futures data that have been ignored or partially ignored by the existing works. 
Firstly, most of the existing methods lack the guidance of financial theory, and are confined to statistical speculation. They ignore various trend signals on price fluctuations and their interactions, which may cause the model to fail to include sufficient explanatory factors, thus limiting the accuracy of the prediction and reducing its applicability in the financial market \cite{cheng2020towards}. 
For instance, Fig. \ref{fig:shortLongTerm} visualizes the 1-minute candlestick price charts as well as the corresponding \{5,10,30\}-minute moving average (MA) trend lines. In actual transaction, we can take the 1-minute candlestick line as short-term signal, the \{5,10\}-minute MA trend lines are intermediate-term signals in different scale while \{30\}- minute MA trend line is corresponding long-term signal. From these MA trend lines, we observe a upside trend from 9:45 to 13:31 start from a cross-over of the 5-minute MA line and 10-minute MA line. This cross-over is usually refer as ``golden cross'' that interpreted by analysts and traders as signaling a definitive upward moment in a market. With this observation, the traders believe it is time to buy with a high probability even still observing jitters in a short-term signals.

Secondly, the price of one future contract usually shows a strong dynamic correlation with other types of future contracts \cite{basak2016model}. This is called cross-sectional interactions. That is, each variable (future contract) depends not only on its historical values (temporal) but also on the observation of other variables.
The cross-sectional spatial relationships contributing to a robust price linkage primarily manifest in the following two forms: 1) Supply and demand relationships along the industrial chain; for instance, a shortage of petroleum may impact the price of synthetic resin in the downstream industrial chain. 2) Substitutable products within the industrial chain, such as the relationship between gold and silver.

This study focuses on modeling multiple futures series by incorporating the two fundamental characteristics simultaneously from the perspective of asset pricing theory. 
The factors that drive asset price changes are highly intricate. Fama and French \cite{fama1993common} proposed a three-factor pricing model by breaking down asset value into the contributions of various measurable factors(market, size, book-to-market), thereby initiating the multi-factor pricing theory. Each factor describes the common exposure of many assets to a certain systemic risk that is the driving force behind the asset's return. Jegadeesh and Titman \cite{jegadeesh1993returns} discovered the momentum factor in the American stock market for the first time. Subsequently, many scholars have provided evidence that momentum effects are widespread in many financial markets around the world. Han \cite{han2013trend} introduced a trend factor related to moving average that surpasses momentum factors in its predictive power. Huang \cite{huang2022momentum} discovered that momentum gap negatively predicts momentum return. Lleo et al. \cite{lleo2023changepoint} evidenced changepoints are a persistent empirical property of stock returns. Inspired by these factor pricing literature, we propose three trend factors that collaboratively explain and forecast future price fluctuations, including a short-term trend factor: gap \cite{huang2022momentum}, an intermediate-term trend factor: moving average (MA) \cite{han2013trend}, and a long-term factor: change-point detection(CPD) \cite{lleo2023changepoint}. At the same time, we propose four heterogeneous tasks, three of which aim to capture the dynamic changes of these trend factors via gap regression, moving average (MA) regression and change-point detection (CPD) respectively, while the final task is a price prediction task. Our rationale is that the gap task captures trend factor that represents the price discrepancy between the maximum and minimum values within a fixed short time interval, which, when combined with the real-time future price forecasting task, can capture short-term features arising from local fluctuations. In contrast, the moving average (MA) task catch the trend factor that depicts the average prices over an intermediate window of time to demonstrate historical volatility of futures prices, allowing for the extraction of intermediate-term features.
The CPD task seize a trend factor that identify abrupt changes in data when a property of the time series changes, while considering the long-term features. Moreover, to prevent "catastrophic forgetting" on short-, intermediate- and long-term features, we require our model to learn an overall experience from all four tasks in a continual manner, as demonstrated in \cite{liu2021overcoming}.
As a whole, a multi-factor predictor framework is constructed by continuous learning of three factor tasks and a forecasting task.

The Spatial Temporal Graph Neural Network (STGNN) \cite{ijcai2018p0505, guo2019aaai, TAN2023109062} extends deep learning methodology to non-Euclidean graphs, making it possible to capture dynamical dependencies. To model the multivariate future data and capture the cross-sectional correlation among futures, we propose to perform STGNNs on the four tasks in a continual learning manner.
When training artificial neural networks (ANNs) on new tasks, previously learned tasks are often forgotten, leading to catastrophic forgetting \cite{liu2021overcoming}. After training on the current task, the model learns a set of optimal parameters that minimize the task's loss. Zenke et al. \cite{zenke2017continual} propose a quadratic surrogate loss to mitigate this issue. The surrogate loss relies on a per-parameter regularization strength, which aims to ``remember" the best model parameters for the last task and adapt to the new task. Existing works \cite{zenke2017continual, liu2021overcoming} approximate the per-parameter regularization strength with the ratio of loss dropping, which is the gradient of the loss.

In our problem, the four tasks are heterogeneous, which results in the corresponding measurement space of their losses being heterogeneous as well. Therefore, using the loss dropping approach based on per-parameter strength is inappropriate for futures price modeling. 
In this paper, we propose a creative solution to substitute the traditional loss change ratio by introducing mutual information between the extracted features and original features. Our rationale is that the mutual information among different tasks is much smoother than their losses. The experimental results demonstrate that our proposed approach achieves higher prediction accuracy and effectively predicts the dynamic change trend of future time series data.

The main achievements, including contributions to the field, can be summarized as follows: 
\begin{itemize}
    \item Conceptually, based on the theory of factor pricing, we construct three trend factors and use heterogeneous tasks to catch the dynamic changes of these factors.
    \item Technically, we propose a novel Heterogeneous Continual Graph Neural Network (HCGNN) to learn temporal (short-, intermediate- and long-term) and spatial dependencies. We design a mutual information-based parameter strength to adapt to heterogeneous tasks.
    \item Empirically, we evaluate the proposed model on a real-world dataset that involves 49 types of futures in the Chinese futures market.
\end{itemize}

The rest of our paper is organized as follows. Section 2 introduces the related work. Section 3 formulates the discussed problem. The components of our proposed Heterogeneous Continual Graph Neural Network(HCGNN) for the multivariate time series forecasting in the futures market are introduced in detail in Section 4. Exhaustive experiments and comparisons with the state-of-the-art methods are shown in Section 5. Section 6 concludes the paper.

\section{Related Work}
\subsection{Deep Learning in Financial Prediction}
A significant amount of literature has focused on applying deep learning in financial prediction. Solving the problem of long-term dependence of the series well, LSTM \cite{hochreiter1997long} and its variants \cite{graves2005framewise, sesti2021integrating, ly2021forecasting} are the preferred choice of most researchers in the field of financial time series forecasting, including but not limited to stock price forecasting\cite{chen2015lstm}\cite{lleo2023changepoint}, index forecasting \cite{baek2018modaugnet,jeong2019improving,siami2018forecasting}, commodity price forecasting \cite{grewal2022machine}, forex price forecasting \cite{ayitey2022forex}, volatility forecasting \cite{zhou2019long}. Salinas et al. \cite{salinas2020deepar} embedded LSTM or its simplified version gated recurrent units (GRU) as a module in the main architecture. Meanwhile, hybrid models combining different deep learning techniques have also been explored. For instance, Wu et al. \cite{wu2021graph} combined CNN (Convolutional Neural Networks) and LSTM to predict stock market movements. They leveraged the ability of convolutional layers to extract relevant financial knowledge as input for LSTM. Combining complementary ensemble empirical mode decomposition (CEEMD), PCA, and LSTM, Zhang et al.\cite{yan2020novel} constructed a deep learning hybrid prediction model for stock markets based on the idea of "decomposition-reconstruction-synthesis". For multivariate time series, Lai et al. \cite{lai2018modeling} introduced a deep learning framework, known as Long- and Short-term Time-series network(LSTNet). LSTNet utilized convolutional neural networks to capture local dependencies. However, LSTNet may not fully exploit latent dependencies between pairs of variables and can be challenging to scale beyond a few thousand-time series. Similar attempts have been made in \cite{shih2019temporal} and \cite{sen2019think}. In general, these methods often completely overlook  available relational information or struggle to fully leverage nonlinear spatio-temporal dependencies.

As a popular model that has emerged in recent years, Graph Neural Networks (GNNs) have gained significant attention from researchers and are widely applicable across various tasks and domains. Due to their exceptional performance in graph modeling, GNNs have found extensive use in the financial field by transforming financial tasks into node classification problems. In order to represent relational data in the financial domain, graphs are commonly constructed. These graphs include user relations graphs \cite{xu2021towards}, company-based relationship graphs \cite{sawhney2020deep}, and stock relationship graphs \cite{li2021modeling, hsu2021fingat}. For instance, Sawhney et al.\cite{sawhney2020deep} focused on stock movement forecasting by blending chaotic multi-modal signals including inter-stock correlations via graph attention networks (GATs) \cite{veličković2018graph}. Recognizing the lead-lag effect in the financial market, Cheng et al. \cite{cheng2022financial} constructed heterogeneous graphs to learn from multi-modal inputs and introduced a multi-modality graph neural network (MAGNN) for ﬁnancial time series prediction. Given the multitude of approaches for constructing graphs, it becomes challenging to determine the most suitable relations for graph construction\cite{wu2020connecting,cao2020spectral}. In addition to constructing graphs from prior knowledge, for multivariate time series forecasting, researchers have also attempted to discover graphical relationships between variables directly from the data. For example, Cao et al. \cite{cao2020spectral} proposed a method called StemGNN, which employed the attention mechanism to construct a graph. They then decomposed the spatio-temporal signal into the spectral domain and frequency domain using Graph Fourier Transform (GFT) and Discrete Fourier Transform (DFT), respectively, for multivariate time-series forecasting. Building upon their work, we extend the application of StemGNN to the context of continual learning.

\subsection{Multi-task Continual Learning}

In the domain of multi-task continual learning (CL), a widely recognized challenge is catastrophic forgetting (CF) \cite{liu2021overcoming}. CF occurs when training a new task causes the model to update existing network parameters, leading to a degradation in accuracy for previously learned tasks. One popular approach to mitigate CF is through regularization, which penalizes changes to important parameters that were learned in the past \cite{zhang2020class}. Another commonly used technique is rehearsal, where examples from previous tasks are stored or generated and replayed during the training of a new task \cite{wang2022memory}. The majority of online CL methods are based on rehearsal. To address the CF issue from a parameter space perspective, Farajtabar et al. \cite{farajtabar2020orthogonal} proposed an approach that restricts the direction of gradient updates to prevent forgetting previously learned data. Another alternative solution for CL is Elastic Weight Consolidation (EWC) \cite{EWC2017pnas} and its variations \cite{pmlrv70zenke17a, NIPS2017_f708f064, Aljundi2018ECCV}. These methods aim to preserve the optimal parameters inferred from previous tasks while optimizing the parameters for subsequent tasks. In the context of GNNs, Liu et al. \cite{liu2021overcoming} extended EWC by introducing a topology-aware weight preserving (TWP) term. However, these models assume that all tasks are isomorphic, sharing the same underlying structure. In this study, we present a novel framework designed specifically for heterogeneous multi-task settings, where tasks exhibit distinct characteristics or have different underlying structures.

\section{Problem Formulation}

\subsection{Preliminary}
In this paper, our focus lies in addressing the problem of multivariate future series forecasting. To effectively capture intra-series and inter-series factors influencing futures prices, we formulate the forecasting problem using a graph structure. This graph can be defined as $\mathcal{G}=(\mathbf{X},\mathbf{A})$, $\mathbf{X}=\left \{ x_{it}  \right \} \in \mathbb{R}^{N\times T}$ are multivariate time-series data, where $N$ is the number of variate (i.e. futures varieties), which corresponds to the number of nodes in the graph $\mathcal{G}$, $T$ denotes the number of time stamps. We define $\mathbf{X}_{t} \in \mathbb{R}^{N}$ as the observed value at timestamp $t$. $\mathbf{A}=[a]_{ij} \in \mathbb{R}^{N\times N} $ is the adjacency matrix of the graph structure, 
where $a_{ij} $ represents the relationship or connection strength between the $i$-th and $j$-th nodes. We utilize a self-attention mechanism to learn the potential correlation matrix $\mathbf{A}$ between multiple futures automatically as in \cite{cao2020spectral}. Hence the strength of the edges in the adjacency matrix can be either positive or negative, allowing for capturing both positive and negative correlations between variables.

\subsection{Four Tasks for Factor Predictor}
\label{sec:4tasks}
\subsubsection{Future Price Forecasting}
The fundamental task is to learn a mapping function $F$ from the historical observations $ \mathbf{X}_{t-T},\dots,\mathbf{X}_{t-2},\mathbf{X}_{t-1} $
to the next $\mathbf{T}^{'}$ timestamps $\mathbf{X}_{t},\mathbf{X}_{t+1},\dots,\mathbf{X}_{t+T^{'}-1}$ ,
\begin{equation}\label{eq:priceForecatingTask}
         [ \mathbf{X}_{t-T},\dots,\mathbf{X}_{t-1} ,\mathbf{A};\Theta,W ] \overset{F}{\rightarrow}[ \mathbf{X}_{t}, \dots, \mathbf{X}_{t+T^{'}-1}  ] ,  
\end{equation}
where $\mathbf{X}_{t} \in \mathbb{R}^{N}$, $\Theta$ and $W$ are the model parameters.

\subsubsection{Three Factor Tracing Tasks }

We propose three carefully designed factor tracing tasks to assist the predictor in capturing trend factors in the data, including 
 \textbf{gap regression}
 , \textbf{moving average (MA) regression}
 , and \textbf{change-point detection (CPD)}

\paragraph{Gap Regression} In the gap regression task, we introduce a measure called the dispersion ratio within a fixed time window as shown in Equation (\ref{eq:gapRregression}) to capture the variation range. The purpose of the gap regression task is to capture short-term trend factor related to local fluctuations.
\begin{equation}
   \Delta \mathbf{X}^{(l)}_t = \frac{ \mathbf{X}_{max} - \mathbf{X}_{min}}{l}
\label{eq:gapRregression}
\end{equation}
where $\mathbf{X}_{max}=\max ( \mathbf{X}[:,t:t+l-1])$ ($\mathbf{X}_{min}=\min ( \mathbf{X}[:,t:t+l-1])$) corresponds to a row maximum (minimum) operation within the sliding window, which returns row maximum (minimum) values of the slice $\mathbf{X}[:,t:t+l-1]$, $l$ denotes the window length.

\paragraph{Moving Average Regression}  The moving average is a type of convolution. It can be viewed as an example of a low-pass filter that can smooth signals. We resort to the moving average to capture the intermediate-term change trends of futures prices. Given a discrete time series with observations $\mathbf{X}_{t}$ , a moving average of $\mathbf{X}_{t}$ can be calculated as follows,
\begin{equation}\label{eq:movingAverage}
   \Bar{\mathbf{X}}_t =  \left ( \mathbf{h}*\mathbf{X} \right )_t = \sum_{i=-M }^{i=M}\mathbf{h}\left ( i \right ) \mathbf{X}\left( t-i \right ) ,
\end{equation}
where $\Bar{\mathbf{X}}_t$ represents the smoothed time series obtained through moving average, $\mathbf{h}$ is a discrete density function (that is $\sum_{i=-M }^{i=M } \mathbf{h}\left ( i \right )=1$), $M$ is the length of the moving average window. The input sequence $\mathbf{X}_{t}$ is the historical settle prices of the futures.


\paragraph{Change-Point Detection} \label{sec:cpd}
Change-point detection is employed in order to provide early warning of long-term trend and estimate the locations where abrupt changes occur within a time series. It is a statistical method that describes the long-term trend.
Given a non-stationary time series $\{ \mathbf{x}_{t}\}_{t=1}^{T}$, a change point refers to the time instance where certain characteristics of the series undergo an abrupt change. Various algorithms exist for change point detection. In this article, we utilize the ruptures library \footnote{https://centre-borelli.github.io/ruptures-docs/}  provided by \cite{truong2020selective} 
to identify the best possible segmentation $\tau $ and the corresponding timestamps or indexes of change points \(\{t_k^*\}_{k=1}^{K}\).  More specifically, we employ a dynamic programming approach, where the objective function is defined as the sum of the costs of all subsequences.
\begin{equation}
 min C(x,\tau)=\sum_{k=1}^{K} c(x_{t_k..t_{k+1}}) 
\label{eq:eq10}
\end{equation}

\begin{equation}
 c(x_{a..b})=\sum_{t=a+1}^{b} {\|x_{t}- \bar x_{a..b}\|}_2^2
\label{eq:eq10}
\end{equation}
here, \(\bar x_{a..b}\) represents the empirical mean of the subsequence \(x_{a..b}, b>a\), where $x_{a..b}:=[x_a, x_{a+1},...,x_{b}]$. The label of change points within the fixed window is defined based on a given threshold $\mathbf \eta (\eta>0)$ as,
\begin{equation}
y_{x_{t_k}}=\left \{
\begin{array}{rcl}
 1 && {max(x_{t_k..t_{k+1}})-x_{t_k}> \eta *x_{t_k} }\\
0 && {max(x_{t_k..t_{k+1}})-x_{t_k} \leq \eta *x_{t_k} } \\
\end{array} \right.
\label{eq:changePointLabel} 
\end{equation}
According to Equation (\ref{eq:changePointLabel}) , when the difference between the maximum value of $x_{t_k..t_{k+1}}$ and $x_{t_k}$ is greater than the threshold $\eta$ times $x_{t_k}$, the change-point factor gives a trend signal of label 1 (going-long). Conversely, when this difference is less than or equal to $\eta$ times $x_{t_k}$, the change-point factor gives a trend signal of label 0 (going short).

It is worth noting that gap regression, moving average regression and change-point detection are suitable for catching short-, intermediate- and long-term trend factors.  
The distinction among short-, intermediate- and long-term trend factors in this paper is relative and depends on the time granularity we set. In addition, regardless of whether it's high-frequency trading (HFT) or low-frequency trading (LFT), the concept of identifying trend factors via factor tasks remains fully applicable.

We believe that the impact of factors on returns cannot be linear, therefore we have avoided using the conventional multi-factor linear regression model. Instead, we have devised four tasks that share data. Each task has its own set of model parameters with constraints, while the other tasks contribute to additional data augmentation for the current learning task, acting as a form of regularization. 

\section{Model}
In this section, we provide a comprehensive formulation of the Heterogeneous Continual Graph Neural Network (HCGNN) based on Mutual Information for factor predictor in futures high-frequency trading. 
\subsection{Overview}
\begin{figure*}[!htp]
    \centering
    \includegraphics[width=0.7\textwidth]{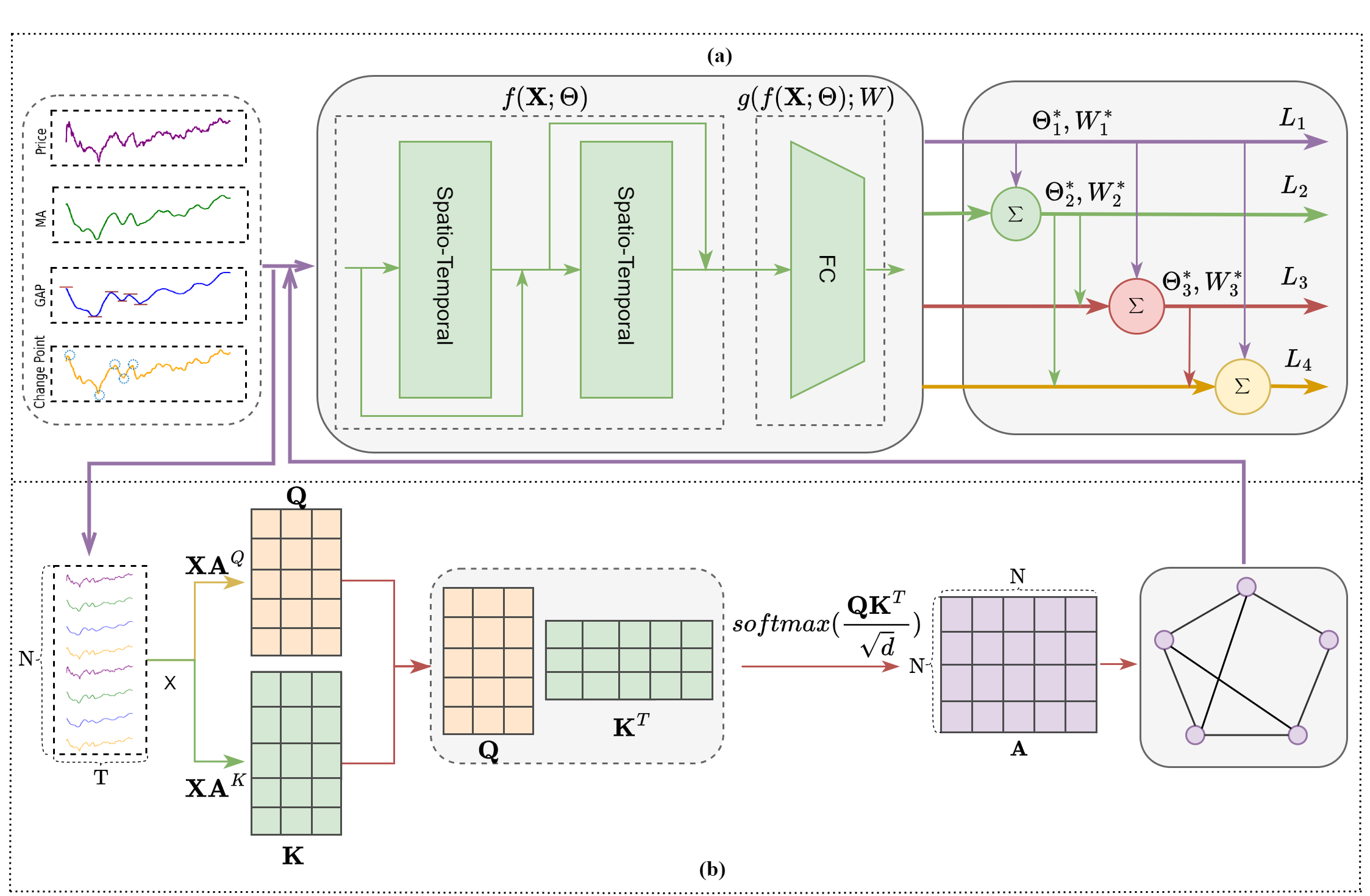}
    \caption{The architecture of the proposed HCGNN is illustrated in Figure 1. Panel (a) depicts the overall framework, where $f(\mathbf{X};\Theta)$ represents a Spatio-Temporal module, and $g(f;W)$ is a feed-forward neural network. In panel (b), the graph construction with self-attention is presented.}
    \label{fig:modelPremeter}
\end{figure*}

We formulate the problem at hand as follows:
\begin{equation}\label{eq:fourTasks}
\left[\mathbf{X}_{t-T},\dots,\mathbf{X}_{t-1},\mathbf{A};\Theta, W\right ]\overset{F}{\rightarrow}
    \begin{cases}
    \left [ \mathbf{X}_{t}, \dots, \mathbf{X}_{t+T^{'}-1} \right ] \\
    \Delta \mathbf{X}^l\\
    \Bar{\mathbf{X}}_t\\
    y(\mathbf{X}_{t_k}), t_k \geq t
    \end{cases}       
\end{equation}
Given the observed values of previous $T$ timestamps $ \mathbf{X}_{t-T},\dots,\mathbf{X}_{t-2},\mathbf{X}_{t-1} $, our objective is to learn an overall mapping function $F:=g(f(x;\Theta); W)$ with parameter $\Theta$ and $W$ as shown in Fig. \ref{fig:modelPremeter}. $f(x;\Theta)$ represents the Spatio-Temporal module with parameter $\Theta$ and $g(f;W)$ is a link function with parameter $W$ that adapts the downstream tasks. Typically, $g$ is implemented as a multi-layer perception or a similar neural network.

Fig. \ref{fig:modelPremeter} depicts the overall framework of the proposed method. Initially, the multivariate futures data $\mathbf{X} \in \mathbb{R} ^{N\times T} $ is inputted into a graph representation layer, which utilizes the self-attention mechanism \cite{vaswani2017attention} to learn the hidden Graph structure $\mathcal{G}$ and correlation weight matrix $\mathbf{A}$ automatically. To calculate the weight matrix $\mathbf{A}$, the following operations are performed,
\begin{equation}
   \mathbf{Q} = \mathbf{X}\mathbf{A}^{Q}, \ \mathbf{K}=\mathbf{X}\mathbf{A}^{T},  \ \mathbf{A} = \text{softmax}\left(\frac{\mathbf{Q}\mathbf{K}^{T} }{\sqrt{d} } \right)
    \label{eq:CorrelationLayer}
\end{equation}
where $\mathbf{X}$ represents the representation of the entire time series, $\mathbf{A}$ is the weight matrix obtained through the self-attention mechanism, $\mathbf{Q}$ and $\mathbf{K}$ denote the query and key representations respectively, which can be computed using linear projections with learnable parameters $\mathbf{A}^{Q}$ and $\mathbf{A}^{T}$ within the attention mechanism. The hidden dimension size of $\mathbf{Q}$ and $\mathbf{K}$ is denoted as $d$. The output matrix $\mathbf{A}\in \mathbb{R} ^{N\times N} $ serves as the adjacency weight matrix for the graph $\mathcal{G} $.

The features $f$ for the four tasks are extracted from the input multivariate futures data using the Spatio-Temporal modules within $f(\mathbf{X};\Theta)$. These modules are interconnected in a residual mode. Subsequently, these features are adapted to the four downstream tasks as shown in Equation (\ref{eq:fourTasks}) with the module $g(f; W)$, which can be a multi-layer perception. The feature extraction process can be represented by the composite function $F:=g(f(\mathbf{X};\Theta); W)$. Finally, we train the four tasks in a continual manner to achieve multi-factor prediction.

Supposing our forecasting task, along with the three trend factor tracing tasks, is denoted as 
$\mathbf{T}:=\{\mathrm{CPD, GAP, MA, PF} \}$. As shown in Equation (\ref{eq:fourTasks}), all four tasks share the feature $\mathbf{X}_{t-T,t-1}$, but their target spaces are heterogeneous. Specifically, $\mathbf{Y}^{1}_t\in \{0,1\}$,  $\mathbf{Y}^{2}_t=\Delta \mathbf{X}^{(l)}_t$, $\mathbf{Y}^{3}_t= \Bar{\mathbf{X}}_t$, $\mathbf{Y}^{4}_t=\mathbf{X}_{t,t+T^{'}-1}$. We train the proposed model in a continual learning manner. Unfortunately, traditional continual learning methods fail to consider a heterogeneous set of tasks. In this work, we propose a mutual information-based method to address this issue which is formulated as follows, 

\begin{equation}\label{eq:continualLearningFrameworkOld}
    \begin{split}
          L_{k} (\Theta,W) & =  L_{k}^{\text{single}} (\Theta,W) \\
          &+
          \lambda_1\sum_{n=1}^{k-1} \sum_{i,j}[\Omega^{\Theta}]_{ij}([\Theta]_{ij}-[\Theta_n^*]_{ij})^2\\
          &+\lambda_2\sum_{n=1}^{k-1} \sum_{i,j} [\Omega^W]_{ij}([W]_{ij}-[W^*_n]_{ij})^2,
    \end{split}
\end{equation}

$\lambda_1$ and $\lambda_2$ are regularization penalties, $L_{k}^{\text{single}}(\Theta, W)$ denotes the loss function of the $k$-th task only, such as cross-entropy loss for classification tasks or mean square error for regression tasks. In contrast, $L_{k} (\Theta,W)$ is the overall loss of all the current $k$ tasks.
$\Theta_n^*$ and $W^*_n$ are the optimal parameters of the $n$-th task. $\odot$ is an element-wise product. $\Omega^{\Theta}$ and $\Omega^W$ serve as ``indicator'' matrices that try to ``divide'' the network parameters $\Theta$ ($W$) into groups for each task. Large values in $\Omega^{\Theta}$ ($\Omega^W$) indicate that the corresponding parts in $\Theta_n^*$ ($W^*_n$) are important memories for the previous task $n$ ($n=0,...,k-1$), hence we need to keep them when training the task $k$. In other words, only the parts of $\Theta$ ($W$) where the indicator values in $\Omega^{\Theta}$ ($\Omega^W$) are small are allowed to be updated. This learning strategy ensures that parameters with low-importance scores can be adjusted freely to adapt to new tasks while penalizing parameters with high-importance scores, thereby enabling the model to maintain good performance on previous tasks.

The crucial part of the continual learning framework is the calculation of the per-parameter regularization strength $\Omega:=\{\Omega^W, \Omega^{\Theta}\}$. 
Generally, $\Omega$ was defined using the gradient of the loss\cite{zenke2017continual,pmlrv70zenke17a}. However, considering the heterogeneous tasks in our problem, we propose to calculate $\Omega$ using mutual information, which will be discussed in Section \ref{sec:Omega}. 

\subsection{Mutual Information Based Per-parameter Regularization Strength $\Omega$}
\label{sec:Omega}
As mentioned earlier, the traditional approach defines the per-parameter regularization strength $\Omega$ as the magnitude of the gradient of loss \cite{zenke2017continual}, 
\begin{equation*}
    \Omega = \left\|  \frac{\partial L}{\partial \Theta} \right \|,
\end{equation*}
however, in our case, the four tasks (as described in Section \ref{sec:4tasks}) are heterogeneous because their targets (Equation (\ref{eq:fourTasks})) lie in different spaces. This implies that the corresponding losses for the four tasks will have different scales. For example, the loss for price regression may differ significantly from the loss for change-point classification. Even for two regression tasks, such as moving average regression and price regression, the losses may lie in different metric spaces. Because the fluctuation of real-time price is more severe than the smooth average.

The heterogeneity of the tasks motivates us to develop new parameter strengths. In this work, we replace the gradient of the training loss with \textit{the gradient of mutual information between the original feature and extracted feature described by the model parameters}.

Specifically, we calculate the mutual information between the original feature $\mathbf{X}^k$ of task $k$ ($k\in\{1, 2, 3, 4\}$) and the feature map $f(\mathbf{X}^{k};\Theta)$ learned from the deep neural network.
For an infinitesimal parameter perturbation on $\Delta \Theta=\{\Delta \boldsymbol\theta_p\}$, the change of mutual information can be approximated as follows,
\begin{equation}\label{eq:motivation1}
    \begin{split}
        &\text{MI}(\mathbf{X}^{k}, f(\mathbf{X}^{k};\Theta + \Delta \Theta)) -  \text{MI}(\mathbf{X}^{k}, f(\mathbf{X}^{k};\Theta))  \\& 
        \approx \sum_m \frac{\partial \text{MI}(\mathbf{X}^{k}, f(\mathbf{X}^{k};\Theta))}{\partial \boldsymbol\theta_p} \Delta \boldsymbol\theta_p,
    \end{split}
\end{equation}
where MI$(x,y)$ represents the mutual information between $x$ and $y$.

From Equation (\ref{eq:motivation1}), we observe that the contribution of the parameter $\boldsymbol \theta_p$ to the change of mutual information can be approximated by the gradient of the mutual information with respect to $\boldsymbol \theta_p$.

Similarly, we can calculate the mutual information between the predicted value $g(\mathbf{Z}^k;W)$ and ground-truth target $\mathbf{X}^k$, where $\mathbf{Z}^k=f(\mathbf{X}^k;\Theta)$ is the feature map output by a deep Spatio-Temporal module as shown in Fig \ref{fig:modelPremeter}. 
Consequently, the change of the mutual information with respect to $\Delta W =\{\Delta \boldsymbol w_q\}$ can be approximated as follows,
\begin{equation}\label{eq:motivation2}
\begin{split}
    &\text{MI}(g(\mathbf{Z}^k; W+\Delta W), \mathbf{X}^k) -  \text{MI}(g(\mathbf{Z}^k;W), \mathbf{X}^k)  \\
    &\approx \sum_m (\frac{\partial \text{MI}(g(\mathbf{Z}^k;W), \mathbf{X}^k)}{\partial \boldsymbol w_q}) \Delta \boldsymbol w_q.
\end{split}
\end{equation}

According to Equation (\ref{eq:motivation1}) and (\ref{eq:motivation2}), we define the per-parameter regularization strength $\Omega:=\{\Omega^W, \Omega^{\Theta}\}$ as follows, 
\begin{eqnarray*}
    \begin{split}
        \Omega^{\Theta} &= \bigg \{ \Big[\frac{\partial [\text{MI}(\mathbf{X}^{k+1} ; \mathbf{Z}^{k+1}) + \text{MI}(g(\mathbf{Z}^{k+1};W) ; \mathbf{X}^{k+1})]}{\partial \boldsymbol\theta_{p}} \Big]_{ij} \bigg\}\\
        \Omega^W &= \bigg\{ \Big[\frac{\partial \text{MI}(g(\mathbf{Z}^{k+1};W) ; \mathbf{X}^{k+1})}{\partial \boldsymbol w_{q}} \Big]_{ij} \bigg \}.
    \end{split}
\end{eqnarray*}

\begin{algorithm}[!htb]
    \caption {Training process of HCGNN. $R$ is a regularization term defined in Equation (\ref{eq:continualLearningFrameworkOld}). }
    \label{alg:model}
    \textbf{Iutput:} $\{\mathbf{X}_{t-T,t-1}, \mathbf{Y}_{t}^{k}\}_{t=T+1}^{L}$ for four tasks $k=1,2,3,4$, $\{\mathrm{1:CPD, 2:GAP, 3:MA, 4:PF} \}$, maximum iteration $M$; \\
    \textbf{Output:} Final model \\
    \begin{algorithmic}[1] 
        \FOR{$epoch$ in  $\text{range}(M)$}  
            \FOR{$(\mathbf{X}_{t},\mathbf{Y}_t^k)$ in $\{\mathbf{X}_{t-T,t-1}, \mathbf{Y}_{t}^{k}\}$}            
                \STATE  $\left \{ \Theta_{1}^{*} , W_{1}^{*}  \right \} \gets \underset{\Theta,W}{\argmin} \left \{ L_{1}:=L_{1}^{\text{single}}(\mathbf{X}_{t},\mathbf{Y}_t^1)  \right \} $
              
                \STATE  $\left \{ \Theta_{2}^{*} , W_{2}^{*}  \right \} \gets \underset{\Theta,W}{\mathrm{argmin} }  \{ L_{2}:=L_{2}^{\text{single}}(\mathbf{X}_{t},\mathbf{Y}_t^2)$ 
                \STATE $\qquad  \qquad  \qquad + R(\Theta_{1}^{*} , W_{1}^{*})   \}$
                
                \STATE $\left \{ \Theta_{3}^{*} , W_{3}^{*}  \right \} \gets \underset{\Theta,W}{\mathrm{argmin} } \{ L_{3}:=L_{3}^{\text{single}}(\mathbf{X}_{t},\mathbf{Y}_t^3)$ 
                \STATE $\qquad  \qquad  \qquad +R(\Theta_{1}^{*} , W_{1}^{*};\Theta_{2}^{*} , W_{2}^{*}) \}$ 
                
                \STATE $\left \{ \Theta_{4}^{*} , W_{4}^{*}  \right \} \gets \underset{\Theta,W}{\mathrm{argmin} } \{ L_{4}:=L_{4}^{\text{single}}({\mathbf{X} _{t}},{\mathbf{Y}_{t}^{4}}) $ 
                \STATE $\qquad  \qquad   \qquad  +R(\Theta_{1}^{*} ,W_{1}^{*}; \Theta_{2}^{*} ,W_{2}^{*}; \Theta_{3}^{*} , W_{3}^{*})  \} $
            \ENDFOR

        \ENDFOR   
        \STATE \textbf{return} $\left \{ \Theta_{4}^{*} , W_{4}^{*}  \right \}$
    \end{algorithmic}
\end{algorithm}

\subsection{The Calculation of $\Omega$}
\label{sec:OmegaC}
In practice, the mutual information defined above is notoriously difficult to calculate. To overcome this problem, it is a popular choice to infer the lower bound and then maximize the tractable objective. van den Oord et al. \cite{Oord2018RepresentationLW} proposed the InfoNCE loss as a proxy objective to maximize the mutual information.
According to their research framework, we assume that $\mathbf{X}^{'} $ is a different view of the input variable $\mathbf{X}$ created through data transformation. We have
\begin{equation}
    \begin{split}
      \max_{\Theta}\text{MI}(\mathbf{X};f(\mathbf{X};\Theta))\ge \max_{\Theta}\text{MI}(f(\mathbf{X};\Theta);f(\mathbf{X}^{'};\Theta))
    \end{split}
\end{equation}

\begin{equation}
    \begin{split}
      \max_{W}\text{MI}(g(\mathbf{Z};W) ; \mathbf{X})\ge \max_{W}\text{MI}(g(\mathbf{Z};W) ; \mathbf{X'})
    \end{split}
\end{equation}

 According to (Oord et al.\cite{Oord2018RepresentationLW}), 
\begin{equation}\label{eq:MIbound1}
    \begin{split}
      \max_{\Theta}\text{MI}(f(\mathbf{X};\Theta);f(\mathbf{X}^{'};\Theta))\ge \log B+\text{InfoNCE}(\left \{\mathbf{X}_{i}\right\}^{B}_{i=1}),
    \end{split}
\end{equation}
and,
\begin{equation}\label{eq:MIbound10}
    \begin{split}
      \text{InfoNCE}(\left \{\mathbf{X}_{i}\right\}^{B}_{i=1})=\frac{1}{B} \sum_{B}^{i=1}log\frac{s(\mathbf{X}_{i},\mathbf{X}^{'}_{i})}{\sum_{j=1}^{B} s(\mathbf{X}_{i},\mathbf{X}^{'}_{j})}, 
    \end{split}
\end{equation}
where $s(\mathbf{X}_{i},\mathbf{X}^{'}_{j})=e^{\frac{f(\mathbf{X};\Theta)^T f(\mathbf{X}^{'};\Theta)}{r} } $ can be regarded as calculating the similarity of $\mathbf{X}_{i}$ and $\mathbf{X}^{'}_{j}$
and $r$ is the temperature. $\{\mathbf{X}_{i}\}^{B}_{i=1}$ are samples from variable $\mathbf{X}$ and $B$ is the batch size. 
We can calculate $\text{MI}(\mathbf{X};f(\mathbf{X};\Theta))$ with Equation (\ref{eq:MIbound1}) and (\ref{eq:MIbound10}). For regression tasks, $\text{MI}(g(\mathbf{Z};W) ; \mathbf{X})$ can be calculated in the same way. However, for classification task, the change point prediction, we have to approximate $\text{MI}(g(\mathbf{Z};W); \mathbf{X})$ as in \cite{guo2022online} that,
\begin{equation*}\label{eq:MIbound2}
    \begin{split}
      &\text{InfoNCE}(\left \{\mathbf{X}_{i}, y(\mathbf{X}_{i}) \right\}^{B}_{i=1})=\alpha \text{InfoNCE}(\left \{\mathbf{X}_{i}\right\}^{B}_{i=1})+\\
      &\beta \sum_{i=1}^B\frac{\sum\limits_{p\in \mathcal{P}} \log \frac{s(\mathbf{X}_{i},\mathbf{X}_{q})s(\mathbf{X}_{i},\mathbf{X}^{'}_{p})s(\mathbf{X}^{'}_{i},\mathbf{X}_{p})}{(\sum_{j=1}^Bs(\mathbf{X}_{i},\mathbf{X}_{j})+s(\mathbf{X}_{i},\mathbf{X}^{'}_{j})+s(\mathbf{X}^{'}_{i},\mathbf{X}_{j}))^3}}{3B\sum_{q=1}^B 1(y(\mathbf{X}_{q})=y(\mathbf{X}_{i}))}
    %
    \end{split}
\end{equation*}
where $\mathcal{P}:=\{p|y(\mathbf{X}_{p})=y(\mathbf{X}_{i})\}$.

\subsection{Spatio-Temporal Module}
As mentioned earlier, we build our model based on Spatio-Temporal Graph Neural Network (STGNN). As the core component of STGNN, the Spatio-Temporal block as shown in Figure \ref{fig:modelPremeter}  is responsible for extracting both spatial and temporal features simultaneously. In this work, we adopt a similar setting as in \cite{cao2020spectral} for the Spatio-temporal feature learning.


\paragraph{Spatial Graph Convolution.}The objective of the spatial module is to extract inter-future features among multivariate time series, i.e., to learn information about the network topology between individual futures varieties. The spatial graph convolution contains three steps: First, projecting the multivariate future series into a spectral domain using the graph Fourier transform (GFT); Second, filtering the outputs from the previous step with a graph convolution (GConv) operator with a learnable kernel; and finally, we apply the inverse graph Fourier transform (IGFT) to produce the final output.

\paragraph{Temporal Sequential Module} The temporal sequential module is primarily responsible for learning intra-series features. It consists of four components: the discrete Fourier transform (DFT), the one-dimensional convolution, the GLU \cite{Dauphin2016LanguageMW}, and the discrete Fourier inverse transform (IDFT). We feed the output of the spatial module to the temporal module as follows. First, we project them into the frequency domain via the DFT; then we apply one-dimensional convolution and subsequent GLU to learn the feature representations in the frequency domain; and finally, the IDFT is adopted to produce the final output.

Following the approach in \cite{cao2020spectral}, we employ multiple Spatio-Temporal blocks with residual connection to create a deeper model. In our experiments, to balance computation cost and model complexity, we utilize two Spatio-temporal blocks. The outputs from both blocks are concatenated and fed into a fully-connected layer to connect each tasks.

\subsection{The Overall Learning Process}
Algorithm \ref{alg:model} summarizes the training process of the proposed model. It is an expanded version of Equation (\ref{eq:continualLearningFrameworkOld}), providing a detailed explanation of the process for continuous learning based on Equation (\ref{eq:continualLearningFrameworkOld}).
$R$ is a regularization term defined in Equation (\ref{eq:continualLearningFrameworkOld}). $L_{k}^{\text{single}}$ is a loss function that depends solely on the data from the $k$-th task.
For example, $L_{1}^{\text{single}}$ is a cross-entropy loss for CPD classification task, $L_{k}^{\text{single}}, k=2,3,4$ are regression losses.

\section{Experimental}
In this section, we aim to evaluate our predictor with other benchmarks on the tick-level high-frequency futures data. 
To gain insights into the effect of different number of factors on the model performance, then we conduct exhaustive ablation studies on the proposed model to validate the necessity to improve the features accordingly.

\subsection{Datasets}
The data used in our experiment is the tick-level high-frequency price series in the Chinese futures market, spanning from 14 February 2022 to 28 February 2022.  The dataset consists of a total of 388,800 price sample points. Given the missing values and newly listed futures varieties without past trading records, we select the most active future contracts as a representative for each type of futures and adopt the following data processing steps,

\begin{enumerate}
    \item Excluding commodity futures more than 50\% of their prices are missing.
    \item Excluding commodity futures with no trading records for the first five days or the last five days.
    \item Excluding commodity futures with missing slot length exceeding  15 minutes within the designated study period. 
    \item Meanwhile, for the extremely rare missing values, we fill them in with the mean. Additionally, all the future series data are normalized using the Z-Score method. 
\end{enumerate}
Following the aforementioned data preprocessing steps, we retain a total of 49 types of commodity futures for further analysis. Subsequently, we split the dataset into three parts for training, validation, and testing with a ratio of 7:2:1, see Table \ref{tab:data_set} for details.

\begin{table}[!t] 
\centering
\caption{Partition of the dataset. \label{tab:data_set}} 
\scalebox{1.0}{
    \begin{tabular}{ccc}
      \hline
        Dataset&Period&samples\\ \midrule
		Training set&02/14/2022-02/23/2022 &278479\\ 
        Validation set&02/23/2022-02/25/2022 &79565\\
        Test set&02/25/2022-02/28/2022& 39782\\ 
      \hline
    \end{tabular}  
    }
\end{table}

\subsection{Experimental Setting}
\subsubsection{Evaluation Metrics} When evaluating the performance of our predictor on the regression tasks, we utilize three commonly used evaluation metrics: select mean Absolute Errors (MAE) \cite{hyndman2006another}, Root Mean Square Errors (RMSE) \cite{hyndman2006another}, Mean Absolute Percentage Errors (MAPE) \cite{hyndman2006another}. To measure the ability of change-point factor tracing task
, we employ precision, recall, accuracy, and F1 score.
To be specific, let $\mathbf{X}^{'}_{t}$ and $\mathbf{X}_{t}$ be the predicted and ground truth values of timestamp $t$, respectively, and $T$ be the total number of timestamps. We define the evaluation metrics as follows,

\begin{itemize}
\item Mean Absolute Error (MAE) \cite{hyndman2006another}:
\end{itemize}

\begin{equation}
 MAE = \frac{1}{T} \sum_{t=1}^{T}\left | \mathbf{X}_{t} - \mathbf{X}^{'}_{t}\right | 
\label{eq:eq10}
\end{equation}

\begin{itemize}
\item Mean Absolute Percentage Error (MAPE) \cite{hyndman2006another}:
\end{itemize}

\begin{equation}
 MAPE = \frac{1}{T} \sum_{t=1}^{T}\left | \frac{\mathbf{X}_{t} - \mathbf{X}^{'}_{t}}{\mathbf{X}_{t}} \right |\times  100 \%  
\label{eq:eq10}
\end{equation}

\begin{itemize}
\item Root Mean Square Error (RMSE)  \cite{hyndman2006another}:
\end{itemize}

\begin{equation}
 RMSE = \sqrt{\frac{1}{T} \sum_{t=1}^{T} (\mathbf{X}_{t} - \mathbf{X}^{'}_{t})^{2} }  
\label{eq:eq10}
\end{equation}

\subsubsection{Baselines} To evaluate the effectiveness of our proposed method, we compare HCGNN against the following baseline methods.
\begin{itemize}
\item \textbf{StemGNN\cite{cao2020spectral}:} Spectral Temporal Graph Neural Network (StemGNN) captures inter-series correlations and temporal dependencies jointly in the spectral domain to further enhance the accuracy of multivariate time-series forecasting.
\item \textbf{MTGNN}\cite{2020Connecting}: Multivariate Time Series Graph Neural Network (MTGNN) utilizes Wavenet as its backbone architecture for temporal modeling of multivariate time series. This model leverages the inherent dependency relationships among multiple series by incorporating a graph learning module. 
\item \textbf{STGCN}\cite{2017Spatio}: Spatio-Temporal Graph Convolutional Networks (STGCN) models traffic flow data by integrating graph convolution
and gated temporal convolution in spatio-temporal convolutional blocks.
\item \textbf{TAMP-S2GCNets}\cite{chen2022tampsgcnets}:
TAMP-S2GCNets integrates time-aware deep learning and multipersistence and constructs a superconvolutional module taking into account both internal and intra-dependencies within the multivariate time series.  
\item
\textbf{SST-GNN}:\cite{sstgnn2021}
Simplified Spatio-temporal Traffic forecasting GNN (SST-GNN) focuses on traffic forecasting. It extracts spatial dependencies by aggregating different neighborhood representations separately and captures temporal dependencies through a weighted spatio-temporal aggregation mechanism.
\end{itemize} 

\subsubsection{Implementation Details}
Using grid search, we get the final setting for the following hyperparameters. Specifically, the dimension of self-attention is set to 32, chosen from a search space of $\{16, 32, 64, 128\}$. The channel size of each graph convolution layer is set to 64 selected from $\{16, 32, 64, 128\}$ and the kernel size of 1D convolution is set to 3, chosen from $\{3, 6, 9, 12\}$. $\lambda_{1}$ and $\lambda_{2}$ are both set to $1\times 10^{-5}$. The sliding window for GAP, MA and CPD is set to 20, 40, 60 respectively. The learning rate is initialized by 0.001 and decayed with a rate of 0.7 after every 5 epochs. The number of training epochs is set as 100. All experiments are conducted on an NVIDIA GeForce RTX 3090 GPU, and the reported results are the best of 20 runs for all models.

\subsection{Overall Performance}
\subsubsection{Results of Forecasting of HCGNN}
\paragraph{Price Forecasting Results}
Table \ref{tab:overall4tasks} presents the main experimental results of 1-minute and 15-minute intervals of the HCGNN model for both the forecasting task and factor tasks. The results indicate that the performance at 1-minute intervals is better than at 15-minute intervals.

Table \ref{tab:overallPerformanceVSbaselines} presents the evaluation against four baselines for price forecasting. The predicted results for both 1-minute and 15-minute time intervals are reported with respect to MAE, RMSE, and MAPE. It is observed that the HCGNN model achieves state-of-the-art results in short-term and relatively long-term price prediction, such as at 1-minute and 15-minute intervals, on this high-frequency trading dataset, which further demonstrates the effectiveness of our modeling approach.

\begin{figure*}[!htp]
    \centering    \includegraphics[width=1.1\textwidth]{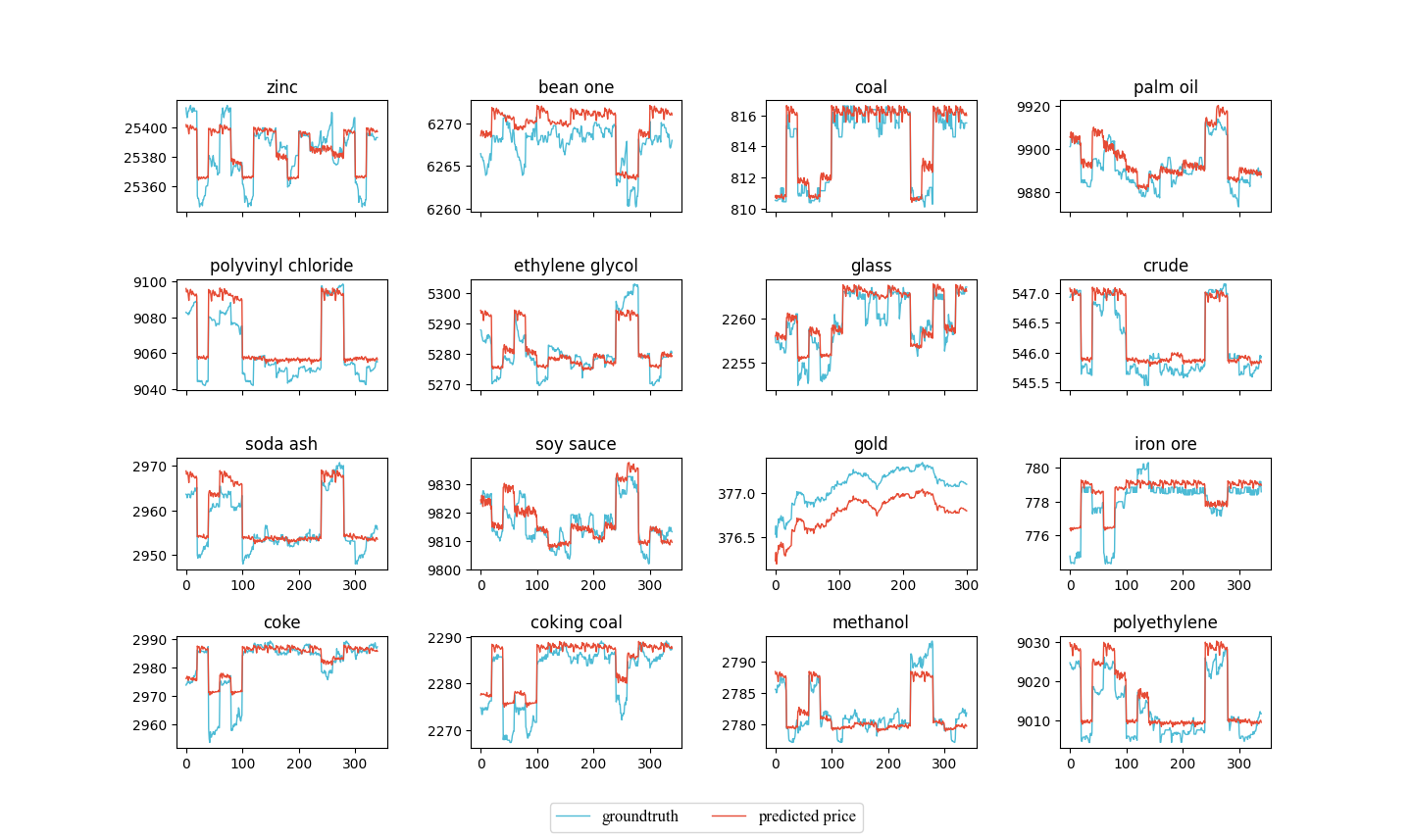}
    \caption{Real-time price forecasting for sixteen representative futures varieties is depicted in the figure. The predicted values are indicated in pink, while the corresponding ground-truth values are presented in blue. The x-axis represents the time in ticks, and the y-axis represents the price of each future variety.}
    \label{fig:visualizationPrice2}
\end{figure*}

We visualize the real-time price prediction of HCGNN for sixteen representative futures varieties, as shown in Fig. \ref{fig:visualizationPrice2}. The predicted values are represented in pink, while the corresponding ground-truth values are shown in blue.

\begin{table}[!t]
\centering
\caption{The overall performance of price forecasting, MA, GAP and CPD with the proposed HCGNN model. Note that we utilize different metrics for regression tasks, such as forecasting, MA, and GAP, as well as for classification tasks CPD.}
\begin{tabular}{c|cccc||cc}
\hline
Interval                 & Metric & Forecast & MA    & GAP  & Metric    & CPD  \\ \hline
\multirow{4}{*}{1min}  & MAE    & 2.52     & 2.03  & 2.02 & Precision & 0.72 \\ 
                       & RMSE   & 7.93     & 6.36  & 6.84 & Recall    & 0.92 \\ 
                       & MAPE   & 0.01     & 0.01  & 0.01 & Accuracy  & 0.73 \\ 
                       &        &          &       &      & F1        & 0.81 \\ \hline\hline
\multirow{4}{*}{15min} & MAE    & 7.54     & 7.43  & 2.78 & Precision & 0.72 \\ 
                       & RMSE   & 19.75    & 18.79 & 9.54 & Recall    & 0.89 \\ 
                       & MAPE   & 0.04     & 0.04  & 0.04 & Accuracy  & 0.72 \\ 
                       &        &          &       &      & F1        & 0.79 \\ \hline
\end{tabular}\label{tab:overall4tasks}
\end{table}

\begin{table}[!t]
\centering
\caption{Comparisons of the performance of close price regression between the state-of-the-art approaches and our method. We report 1-minute and 15-minute results.} \label{tab:performance_comparison}
\resizebox{\columnwidth}{!}{         
\begin{tabular}{ccccccc}
\hline
\multirow{2}{*}{Models}&
\multicolumn{3}{c}{1min}&\multicolumn{3}{c}{15min}\cr
\cmidrule(lr){2-4} \cmidrule(lr){5-7} 
&MAE&RMSE&MAPE(\%)  &MAE&RMSE&MAPE(\%)\cr
\midrule
StemGNN  &15.24     &52.35     &0.08    &21.63     &73.69     &0.16         \cr       
		MTGNN &64.58     &126.08     & 0.77 & 84.29    & 171.09    & 1.19              \cr
        STGCN	 & 16.79     & 36.88  & 0.19   & 16.94     & 37.23  & 0.23 \cr
        SSTGNN	 & 68.68    & 135.75  & 1.09    & 223.80    & 389.30  & 4.34 \cr 
        HCGNN	&{\bf 2.52}&{\bf 7.93}&{\bf 0.01}&{\bf 7.54}&{\bf 19.75}&{\bf 0.04}\cr
\hline
\end{tabular} } 
\label{tab:overallPerformanceVSbaselines}
\end{table}

\subsubsection{Portfolio with HCGNN}
In this section, we aim to validate the model's performance by conducting an investment simulation using the classic top-$K$ trading strategy. The simulation is based on signals provided by the HCGNN model as well as other baseline methods separately. To be more specific, we assume a trading cost of 0.001 and implement the top-$K$ ($K=10$) trading strategy by investing in the 10 highest increase or decrease in the 2-minute forecasted price compared to the current price.
In this simulation, the budget is evenly distributed among the top 10 contracts, both selling short and buying long were allowed. Specifically, if the 2-minute forecasting price is going up and the increasing percentage is greater than 0.1\%, then we open the long position on the contract immediately and closed it 2 minute later. Otherwise, if the 2-minute forecasted price is going down and the decreasing percentage is greater than 0.001, then we open the short position on the contract immediately and closed it 2 minute later. 
Meanwhile, we re-balance the portfolio every two minutes.

\begin{figure}[!t]
    \centering    \includegraphics[width=\columnwidth]{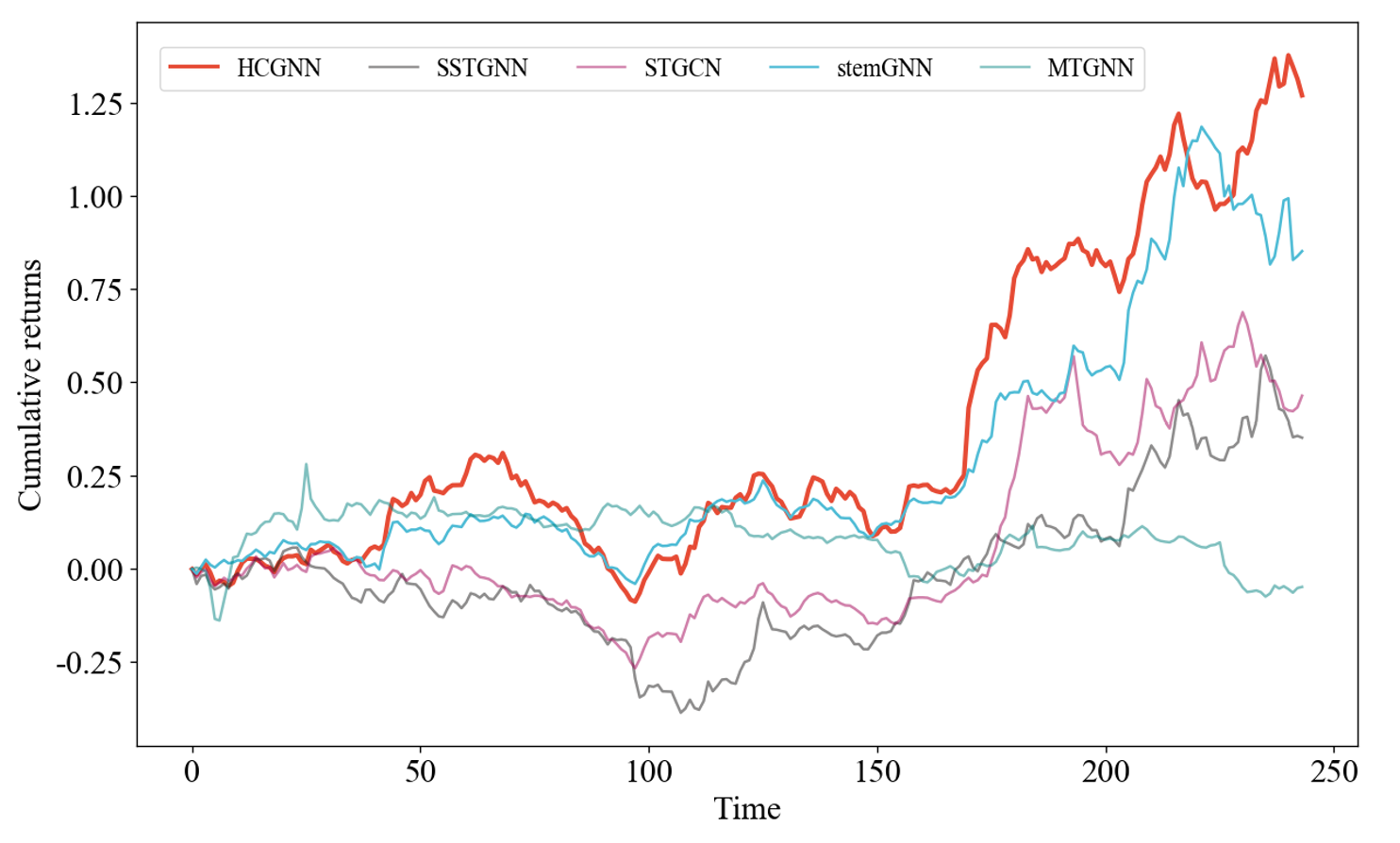}
    \caption{Comparisons of cumulative return curves with the prediction results of four baselines and the proposed HCGNN. }
    \label{fig:portfolio}
\end{figure}

 We use the test set data ranging from  February 25 to 28, 2022 for backtesting. Fig. \ref{fig:portfolio} presents the accumulated minute return (y-axis) using the top-N trading strategy based on the HCGNN and other baseline methods over the assessment period (x-axis). The cumulative return is the difference between the closing profit and a 0.1\% trading cost.

Fig. \ref{fig:portfolio} demonstrates that among all methods, HCGNN stands out with a remarkable return up to 140\%, which is significantly higher than other methods. In contrast, the MTGNN approach ends the evaluation period with an underwhelming return of -4.8\%. This result further validates the exceptional performance of the proposed predictor, which is uniquely advantaged by its effective integration of multi-factor pricing mechanism and the future correlationship consideration. Notably, this profit was derived from both long and short positions. The profit generated through long positions amounted to 60.51\%, whereas the profit from short positions was 73.84\%. Selling short emerged as a more profitable strategy during the evaluation period, as the overall market trend was downward.

\subsection{Visualization of the Three Factors}

\subsubsection{Change Points Prediction}
\begin{figure*}[!t]
    \centering    \includegraphics[width=1.0\linewidth]{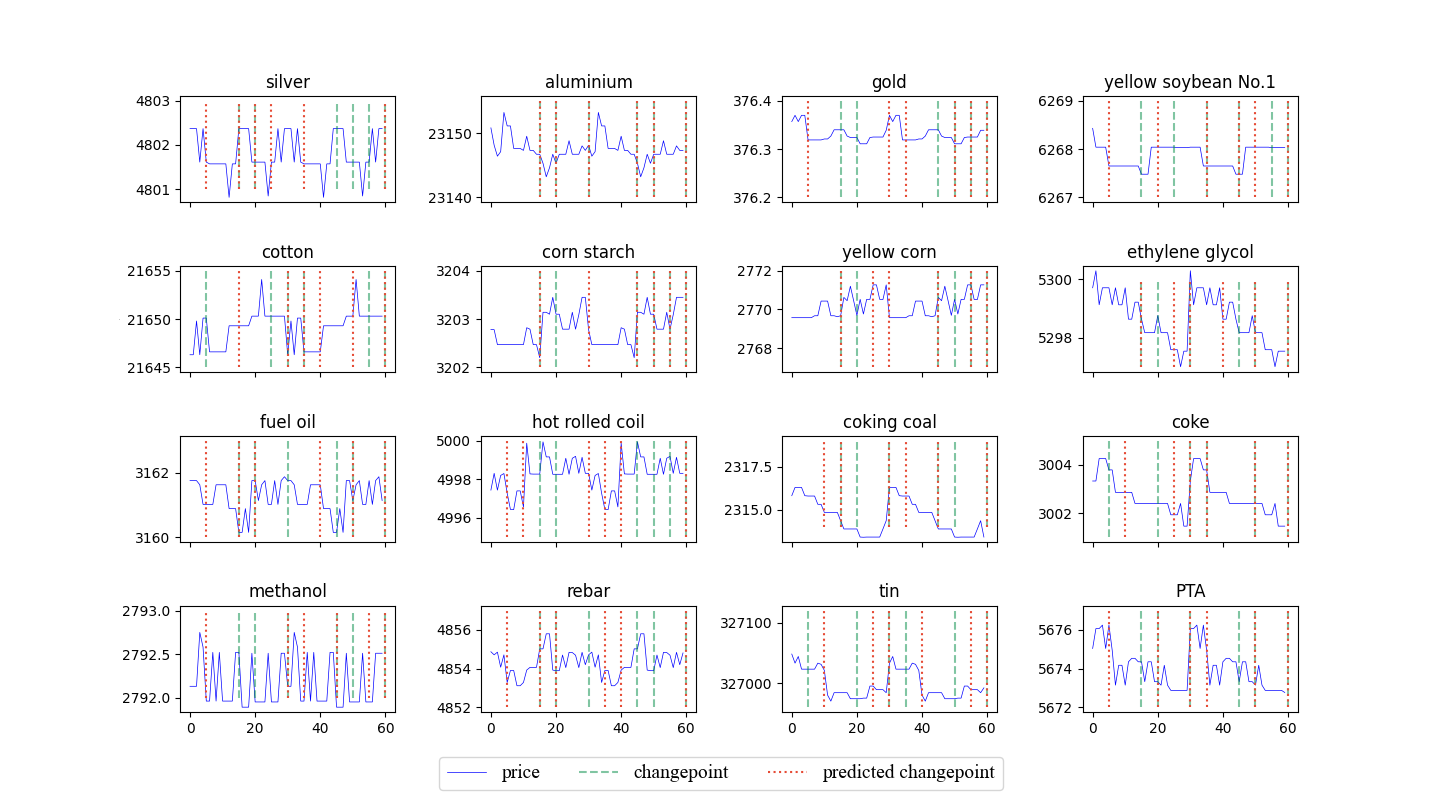}
    \caption{Change point detection visualization. The pink solid curve is the future price. The vertical dashed red line and dashed green line represent the real change points and predicted change points respectively. The x-axis denotes the time slot in tick and the y-axis denotes the price of each variety of future.}    \label{fig:visualizationPrice_cpd}
\end{figure*}

It is crucial for investors to detect changes in trading trends or trend reversals in a timely, early, and accurate manner, whether they are subjective traders or quantitative traders, involved in high-frequency trading (HFT) or low-frequency trading (LFT). In Figure \ref{fig:visualizationPrice_cpd}, the change-point factor tracing of sixteen varieties of future contracts is presented. Vertical lines are used to mark the change points, where green lines represent the real change points and red lines indicate the predicted change points. The ground truth is obtained using a change point detection algorithm, as detailed in Section \ref{sec:cpd}. From the figure, it is evident that most of the vertical dashed red lines (representing the predicted change points) overlap with the green lines (representing the ground truth), suggesting that the model can successfully detect the points at which abrupt changes occur in the trading trends. Furthermore, the label values associated with these change points can provide insights into whether the price will rise or fall.

\subsubsection{Moving Average}
The moving average is a commonly used trend factor in technical analysis that provides a smoothed representation of the price over a specified period. 
Unlike the raw price curve, which can exhibit significant fluctuations, the moving average curve is smoother and reveals the intermediate-term market movements in this article, as shown in Fig. \ref{fig:shortLongTerm}. 
   
\begin{figure}[!t]
    \centering
    \includegraphics[width=1.0\columnwidth]{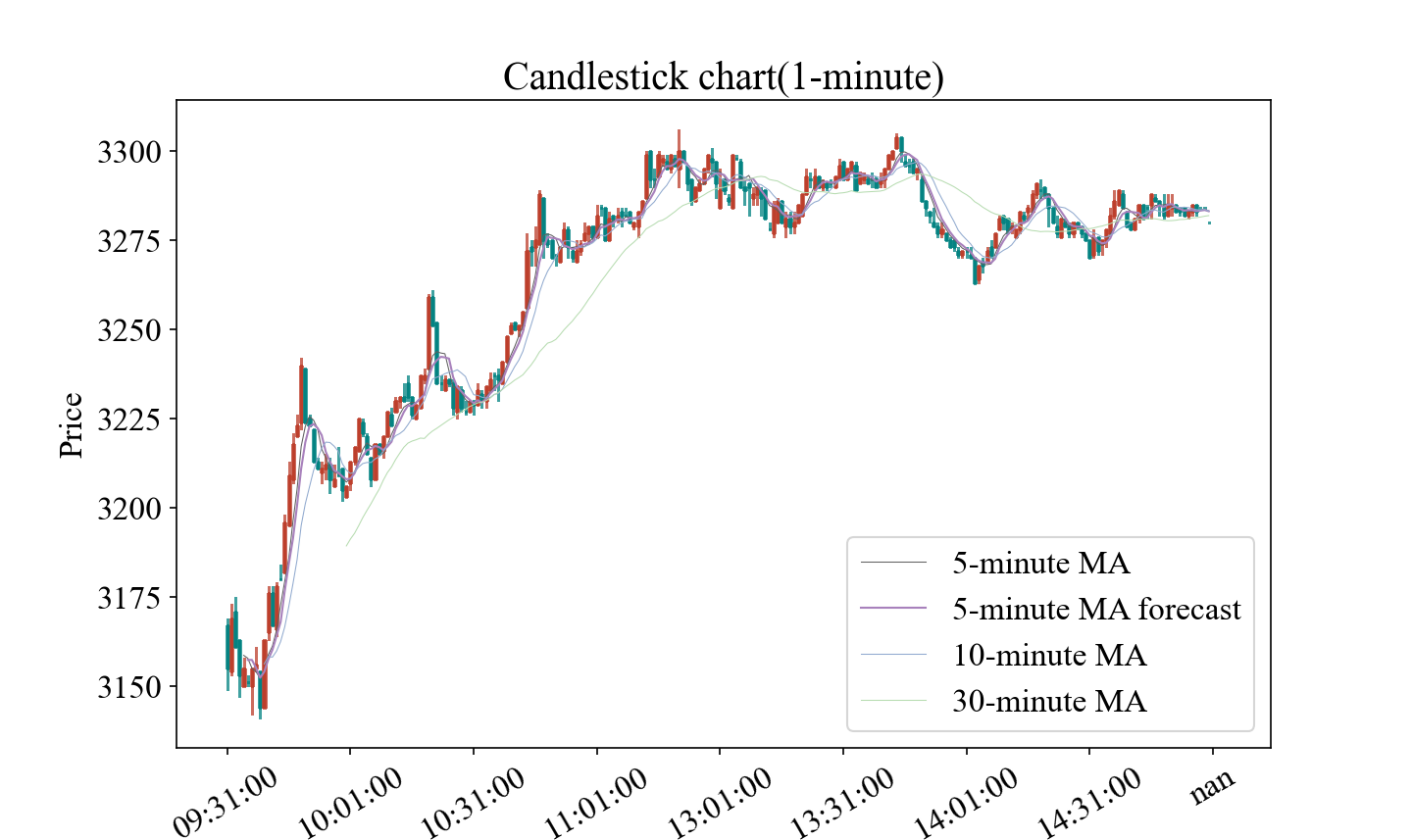}
    \caption{Moving Average lines. The top half chart illustrates the 1-minute k-line. The lower left and lower right parts show 5-minute and 15-minute k-line.}
    \label{fig:visualizationMA}
\end{figure}

\subsubsection{Gap Regression} 
\begin{figure}[!t]
    \centering
    \includegraphics[width=1.1\columnwidth]{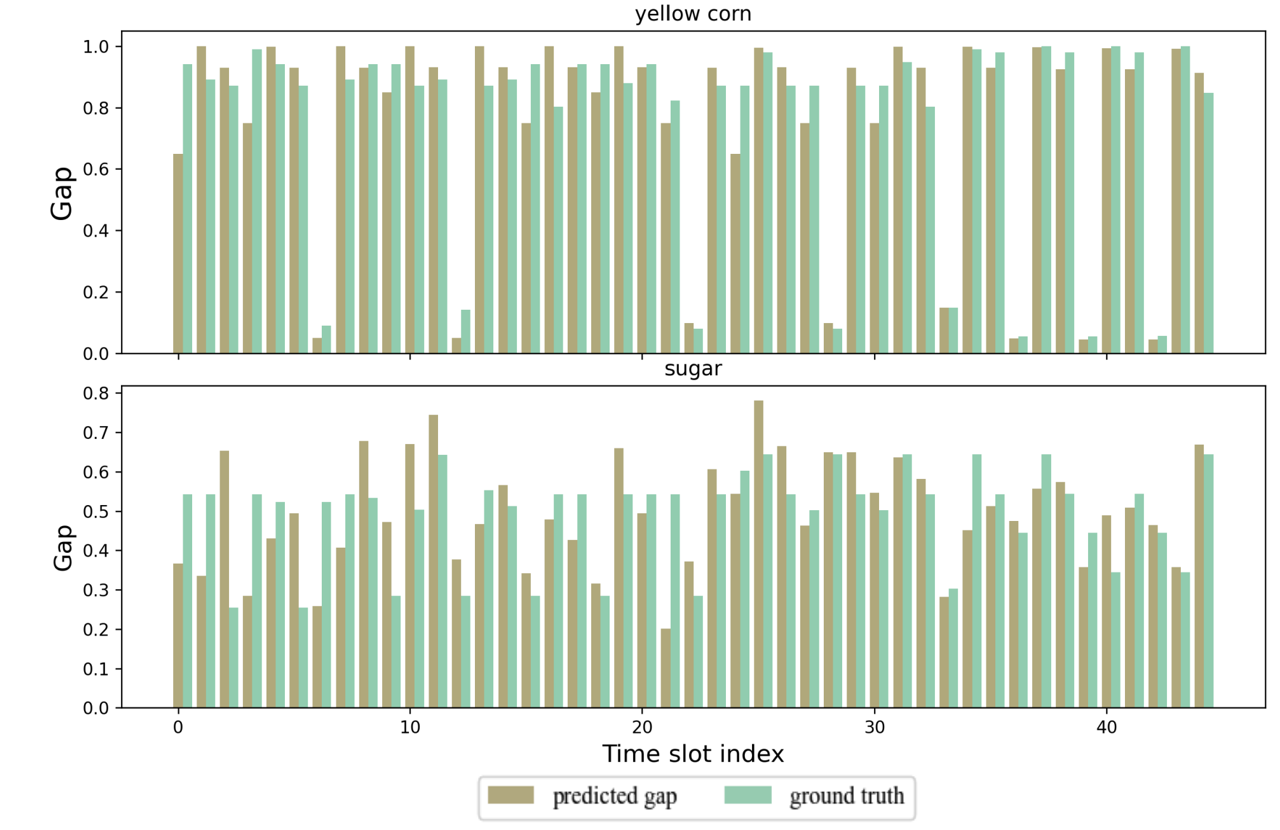}
    \caption{Gap regression. The orange bar and blue bar represent the ground truth and the predicted gap respectively.}
    \label{fig:visualization_gap}
\end{figure}

The gap factor task in our model depicts the dispersion degree of futures prices. Fig. \ref{fig:visualization_gap} shows a visualization of the regression results of corn and sugar contracts. The brown bars are the ground truth, indicating the actual price gap in a window of time. The blue bars represent the predicted gap, which is the model's estimation of the dispersion degree. The predicted value is close to the ground truth, indicating that the model is effective in estimating the dispersion degree for the time series. 

\subsection{Ablation Studies}
In our HCGNN model, there are three heterogeneous factor tasks. To verify whether the designed factors provide additional explanatory and predictive power and to evaluate the effectiveness of each task in the HCGNN model, we designed several variants of the model and conducted an ablation study using high-frequency futures data. We refer to the HCGNN model without different components as follows, and the results on futures data are presented in Table \ref{tab:ablation_study}:
\begin{itemize}
\item\textbf{w/o GAP+CPD+MA}: HCGNN without the GAP, CPD, and MA factor tasks. We train the model only on the single forecasting task.
\item\textbf{w/o GAP+CPD}: HCGNN without the GAP and CPD factor tasks. We train the model on the Forecasting and MA factor tasks in continuous learning with mutual information.  
\item\textbf{w/o MA+CPD}: HCGNN without the MA and CPD factor tasks. We train the model on forecasting and gap factor tasks in continuous learning with mutual information.
\item\textbf{w/o GAP+MA}: HCGNN without the gap and MA factor tasks. We train the model on forecasting and CPD factor tasks in continuous learning with mutual information.
\item\textbf{w/o MA}: HCGNN without the MA factor task. We train the model on forecasting, gap, and CPD factor tasks in continuous learning with mutual information.
\item\textbf{w/o GAP}: HCGNN without the gap factor task. We train the model on forecasting, MA, and CPD factor tasks in continuous learning with mutual information. 
\item\textbf{w/o CPD}: HCGNN without the CPD factor task. We train the model on forecasting, MA, and gap factor task in continuous learning with mutual information.
\item\textbf{w/o MI}: HCGNN without mutual information. We train the model on all four tasks continuously but replace mutual information with the traditional gradient of loss. 
\end{itemize}

For all the ablation experiments, we only back-propagate the gradients of the activated tasks. Take the setting \textbf{w/o GAP+CPD+MA} as an example, we set the \texttt{requires\_gradient} as \texttt{True} for the price \textbf{forecasting} task while restraining the gradients of GAP, CPD, and MA.

\begin{table*}[!t] 
  \centering
  \fontsize{8}{12}\selectfont   
  \caption{Ablation study on price regression.}      \label{tab:ablation_study}
       \scalebox{1.0}{
        \begin{tabular}{cccccccccccc}
        \hline
        \rule{0pt}{12pt}
        \multirow{2}{*}{Tasks}&
        \multicolumn{3}{c}{1min}&\multicolumn{3}{c}{15min} 		\cr
        \cmidrule(lr){2-4} \cmidrule(lr){5-7} \cmidrule(lr){8-11} 		
        &MAE&RMSE&MAPE(\%) &MAE&RMSE& MAPE(\%) 
		\cr
        \midrule
          w/o GAP+CPD+MA&4.1820&13.1008&0.0278  &11.5928&36.9416&0.0933\cr
        w/o GAP+CPD&3.3287&9.4018&0.0236&9.9714&32.6971 &0.0768\cr
        w/o MA+CPD&3.5683&9.5595&0.0253&10.4207&33.1246 &0.0774\cr
        w/o GAP+MA&3.7432&9.5364&0.0229&11.0984&33.9850 &0.0816\cr
        w/o MA&2.9518 &9.2294 &0.0211 &9.1012 &26.9308 &0.0535\cr
		w/o GAP&2.8674 &9.1361 &0.0149 &8.6386 &25.8250 &0.0588\cr
		w/o CPD&2.8638 &9.1593 &0.0188 &8.6637 &25.7559 &0.0504\cr  
        w/o MI&6.8644&17.6328&0.1236&16.6115&48.7536&0.2857\cr
        {\bf HCGNN}	&{\bf 2.5161}&{\bf 7.9261}&{\bf 0.0126}&{\bf 7.5356}&{\bf 19.7487}&{\bf 0.0409}\cr
	    \hline
        \end{tabular}
        }
\end{table*}

Based on the ablation experimental design approach described above, we evaluated the various variants described above on the high-frequency trading dataset and present all the experimental results in Table \ref{tab:ablation_study}. Values in bold are the best results of these settings. We have several observations based on the results of 1-minute and 15-minute:
\begin{itemize}
\item The performance of the model improves as the number of factor tasks increases, the best forecasting performance is achieved when all three factor tasks are engaged simultaneously with the forecasting task. This indicates that the three trend factors carefully designed collectively capture the short-, intermediate-, and long-term trend risks and can provide increments for price prediction. From the perspective of machine learning, it also tells us learning different supervised labels through multi-task learning is beneficial for improving prediction performance \cite{cheng2020towards}.
\item Mutual information is obviously a more suitable method for defining parameter strength in the heterogeneous multi-task setting compared to the traditional loss-based methods.
\item The performance of the CPD factor task slightly declines in the HCGNN model which indicates it is challenging to improve the performance of all tasks in multi-task continual learning. 
\end{itemize}

\subsection{Results Analysis}

\subsubsection{Learning Curves of HCGNN} 
\begin{figure}[!t]
    \centering    \includegraphics[width=1.0\columnwidth]{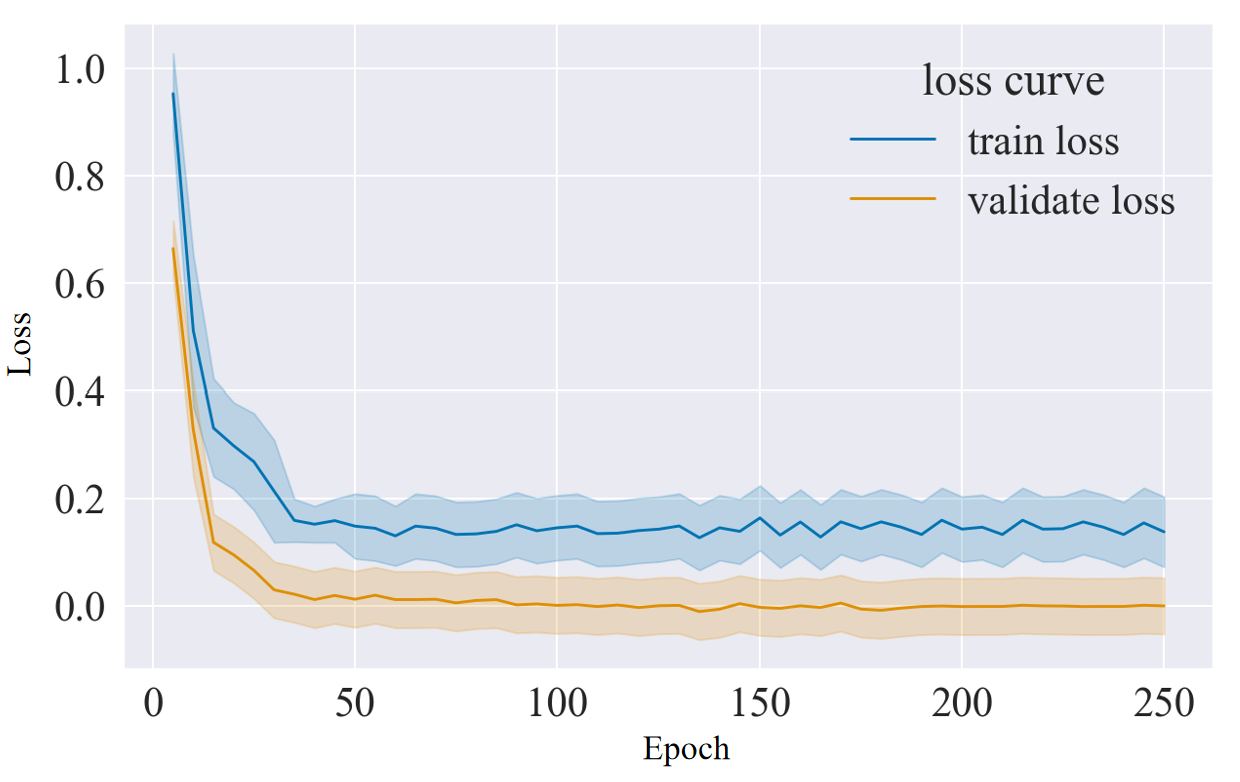}
    \caption{Loss variation in training and validation.}
    \label{fig:my_label_loss}
\end{figure}

The learning curves of HCGNN on the high-frequency futures data are illustrated in Fig. \ref{fig:my_label_loss}, where the x-axis represents the time slot and the y-axis represents the total loss for the forecasting task. As depicted in the figure, we observe the convergence of the HCGNN model when trained for 100 epochs.

\subsubsection{Correlation between Futures Varieties} 
\begin{figure}[!t]
    \centering    \includegraphics[width=1.0\columnwidth]{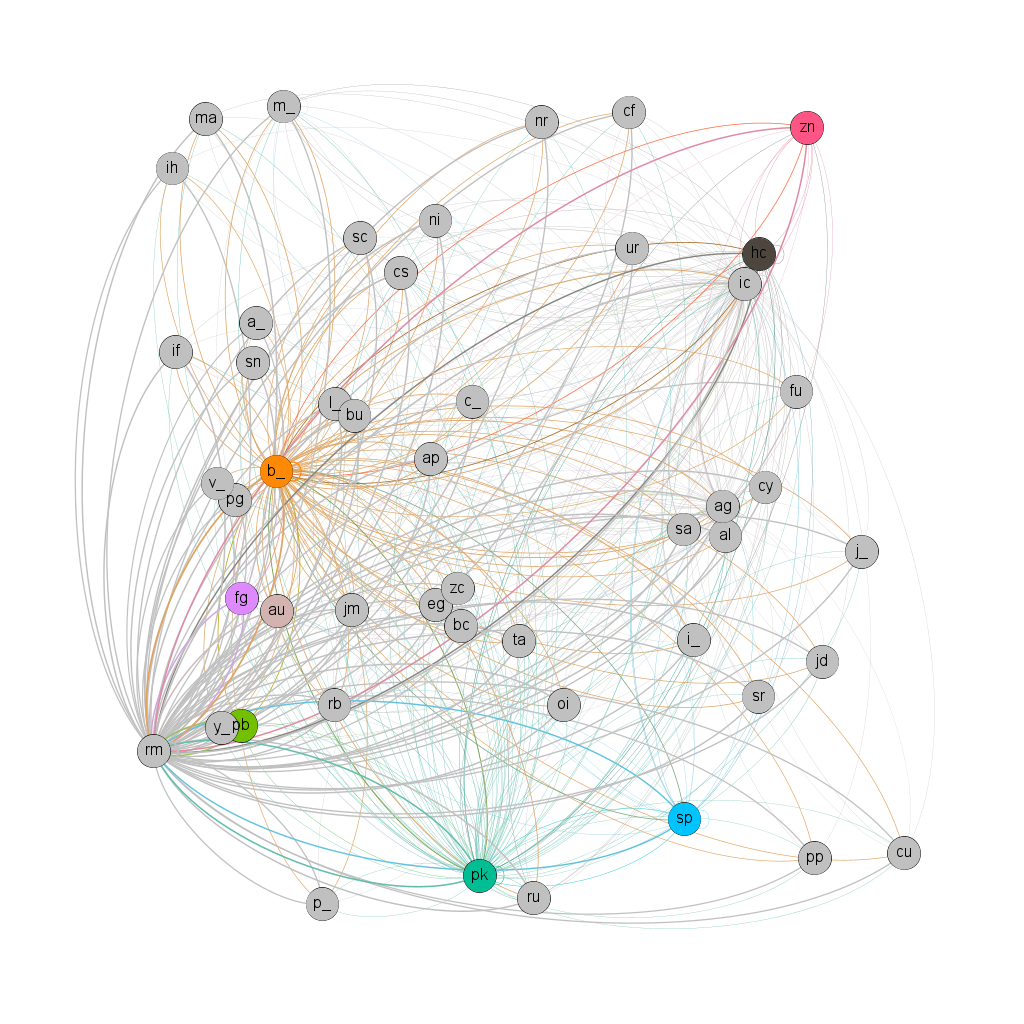}
    \caption{Correlation between Futures Varieties. }
    \label{fig:my_label_top}
\end{figure}

To model the correlation between futures varieties, we employ an automated approach to construct a graph. The adjacency matrix of the graph represents the correlation strength between futures varieties. As visualized in Fig. \ref{fig:my_label_top} in the form of network topology, the thicker the line between two nodes, the stronger the correlation between the two varieties of futures. Simultaneously,  the density of connections among nodes reflects the influence of each future variety. 
The abbreviations and full names of futures varieties are provided in Table \ref{tab:abb}. From Fig. \ref{fig:my_label_top}, several observations can be made:
\begin{itemize}
    \item First, as we can see, futures varieties belonging to a complete upstream and downstream industrial chain exhibit strong linkages. This includes varieties such as bean futures varieties (bean one(a\underline{~}), bean two(b\underline{~}), soybean meal(m\underline{~}), rapeseed meal(rm), and soy sauce(y\underline{~}), etc), black futures (rebar(rb), hot rolled coil(hc), coke(j\underline{~}), coking coal(jm), iron ore(i\underline{~}), coal(zc), etc), crude(sc) and crude refined products (polyethylene (L\underline{~}), polypropylene(PP), PTA(ta), polyvinyl chloride(v\underline{~}),  bitumen(bu), etc).
    \item Second, commodities that have a substitution relationship (eg. gold(au) and silver(ag); soybean meal(m\underline{~}), rapeseed meal(rm), palm oil(p\underline{~})) are closely related to each other. 
\end{itemize}

\begin{table*}[!t] 
\centering
\caption{Futures variety and its abbreviation\label{tab:abb}} 
    \begin{tabular}{cccccc}
      \hline
        ABB&Full Name&ABB&Full Name&ABB&Full Name\\ \midrule
	AG&Silver&AL&Aluminium&AP&Apple\\ 
    AU&Gold&A&Bean One&BC&International Copper\\ 
    BU&Bitumen&B&Bean Two&CF&cotton\\ 
    CS&Cornstarch&CU&Shanghai Copper&CY&Cotton Yarn\\
    C&Yellow Corn&EG&Ethylene Glycol &FG&Glass\\
FU&Fuel Oil&HC&Hot Rolled Coil\\
IC&CSI 500 Stock Index&IF&CSI 300 Stock Index\\
IH&\multicolumn{2}{c}Shanghai Stock Exchange 50 Stock Index\\
I&Iron Ore&JD&Egg&JM&Coking Coal\\
J&Coke&L&Polyethylene&MA&Methanol\\
M&Soybean Meal&NI&Nickel&NR&International rubber\\
OI&Vegetable Oil&PB&lead&PG&liquefied petroleum gas\\
PK&Peanut&PP&Polypropylene&P&Palm Oil\\
RB&Rebar&RM&Rapeseed Meal&RU&Domestic Rubber\\
SA&Soda Ash&SC&Crude&SN&tin\\
SP&Pulp&SR&Sugar&TA&PTA\\
UR&Urea&V&Polyvinyl Chloride&Y&soy sauce\\
ZC&Coal&ZN&Zinc\\
      \hline
    \end{tabular}  
  
\end{table*}
\section{Conclusion and Future Work} 
In this work, we introduce a novel factor predictor called Heterogeneous Continual Graph Neural Network (HCGNN). It models spatio-temporal futures data and captures the correlations between different varieties via graph learning. 
Inspired by factor pricing theory, we design three factor tasks and prediction task to learn in a continual manner to construct a factor predictor. Furthermore, we creatively make use of a mutual information maximization mechanism to address the challenge of CF in continuous learning. Experimental results demonstrate that HCGNN consistently outperforms existing approaches in future time-series forecasting. As for future research directions, we aim to explore the applicability of HCGNN in stock, bond, forex, option, and other financial markets.

{\appendix[Visualizations of gap predictions on more futures]
\setcounter{table}{0} 
\setcounter{figure}{0} 
\setcounter{equation}{0} 
\renewcommand{\thetable}{\thesection-\arabic{table}} 
\renewcommand{\theequation}{\thesection-\arabic{equation}}
More visualizations of gap predictions are as shown in Fig. \ref{fig:gapRegression}.
\begin{figure*}[!t]
\centering    
\includegraphics[width=\textwidth]{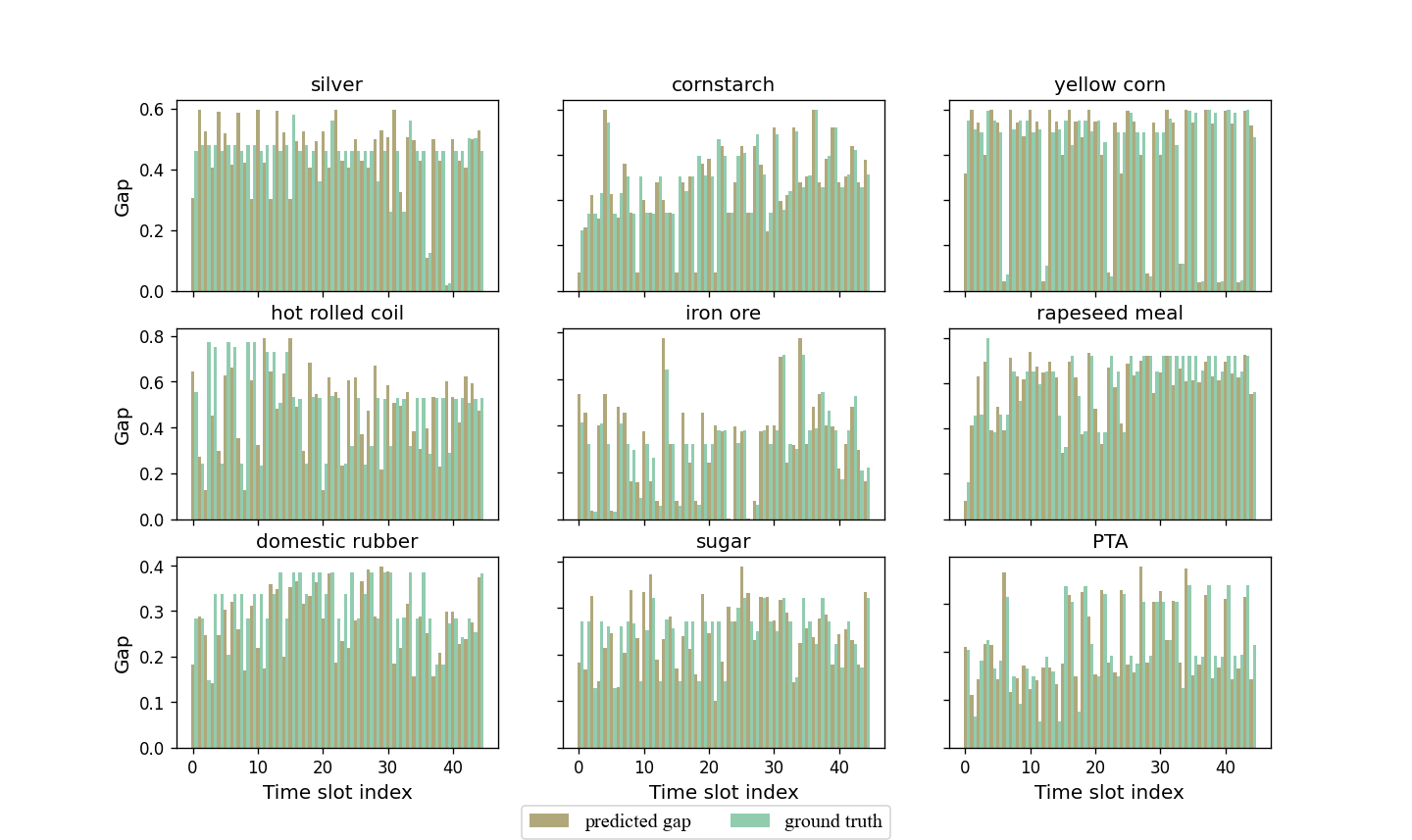}
\caption{Visualizations of gap predictions on more futures. Predicted gap and the corresponding ground truth are plotted in blue and brown bars respectively.}
\label{fig:gapRegression}
\end{figure*}

}

\bibliographystyle{IEEEtran}
\bibliography{ref}

\end{document}